\documentclass[sigconf]{acmart}

\usepackage{caption,tabularx,booktabs, arydshln}
\usepackage{xcolor}
\usepackage{comment}
\usepackage{multirow}
\usepackage{adjustbox}
\usepackage{rotating}
\usepackage{enumitem} 
\usepackage{utfsym}
\usepackage{hhline}
\usepackage{colortbl}
\AtBeginDocument{%
  \providecommand\BibTeX{{%
    \normalfont B\kern-0.5em{\scshape i\kern-0.25em b}\kern-0.8em\TeX}}}

\begin{document}

\title{REACT 2026: The Fourth Multiple Appropriate Facial Reaction Generation Challenge: Personalised MAFRG and Appropriate EEG Reaction Prediction}

\author{Siyang Song}
\authornote{Corresponding Author}
\affiliation{%
  \institution{University of Exeter}
\country{United Kingdom}
  }
  \email{s.song@exeter.ac.uk}

\author{Micol Spitale}
\affiliation{%
  \institution{Politecnico di Milano}
  \city{Milan}
  \country{Italy}
  }
  \email{micol.spitale@polimi.it}

\author{Zijian Wu}
\affiliation{%
  \institution{Nanjing University of Science and Technology}
  \country{China}
  }
  \email{wuzijian@njust.edu.cn}

\author{Xiangyu Kong}
\affiliation{%
  \institution{University of Exeter}
  \country{United Kingdom}
  }
  \email{xk219@exeter.ac.uk}


\author{Cheng Luo}
\affiliation{%
  \institution{King Abdullah University of Science and Technology}
  \country{Saudi Arabia}
  }
  \email{cheng.luo@kaust.edu.sa}

  \author{Cristina Palmero}
\affiliation{%
  \institution{King's College London}
  \city{London}
  \country{United Kingdom}}
\email{cristina.palmero@kcl.ac.uk}

  \author{German Barquero}
\affiliation{%
  \institution{Universitat de Barcelona}
  \city{Barcelona}
  \country{Spain}}
\email{germanbarquero@ub.edu}

  \author{Sergio Escalera}
\affiliation{%
  \institution{Universitat de Barcelona}
  \city{Barcelona}
  \country{Spain}}
\email{sergio@maia.ub.es}

  \author{Michel Valstar}
\affiliation{%
  \institution{University of Nottingham}
  \city{Nottingham}
  \country{United Kingdom}}
\email{michel.valstar1@nottingham.ac.uk}

  \author{Mohamed Daoudi}
\affiliation{%
  \institution{IMT Nord Europe}
  \city{Villeneuve d'Ascq}
  \country{France}}
\email{mohamed.daoudi@imt-nord-europe.fr}


  \author{Fabien Ringeval}
\affiliation{%
  \institution{Université Grenoble Alpes}
  \city{Grenoble}
  \country{France}}
\email{fabien.ringeval@imag.fr}

\author{Andrew Howes}
\affiliation{%
  \institution{University of Exeter}
  \city{Exeter}
  \country{United Kingdom}}
\email{andrew.howes@exeter.ac.uk}

\author{Elisabeth Andrè}
\affiliation{%
  \institution{University of Augsburg}
  \city{Augsburg}
  \country{Germany}}
\email{andre@uni-a.de}

\author{Hatice Gunes}
\affiliation{%
  \institution{University of Cambridge}
  \city{Cambridge}
  \country{United Kingdom}}
\email{hatice.gunes@cl.cam.ac.uk}

\renewcommand{\shortauthors}{Song et al.}

\begin{abstract}

In dyadic interactions, various human facial reactions could be \textit{appropriate} for responding to each human speaker behaviour. Following the successful organisation of the REACT 2023, 2024 and 2025 challenge series, a body of generative deep learning (DL) models have been developed for the problem of multiple appropriate facial reaction generation (MAFRG). This year, we propose the REACT 2026 challenge encouraging the development and benchmarking of Machine Learning (ML) models that can generate multiple \textbf{personalised}, appropriate, diverse, realistic and synchronised human-style facial reactions expressed by a specific human listener for responding to each given speaker behaviour. As a key of the challenge, we continuously provide challenge participants with MARS dataset introduced by REACT 2025 but \textbf{additionally provide individual-level Big-Five personality labels and EEG recordings}. This introduces a new one-to-many personalised facial reaction generation setting combining human expressive behavioural, affective and neurophysiological signals, which remains largely unexplored in current dyadic interaction modelling. This paper also presents the challenge guidelines and new baselines on the four proposed sub-challenges: Offline generic and personalised MAFRG as well as Online generic and personalised MAFRG, respectively, which are publicly available at \url{https://github.com/reactmultimodalchallenge/baseline_react2026}.

\end{abstract}





\maketitle


\section{INTRODUCTION}



Personalised human-style facial behaviours play a key role for people to convey their unique characteristics, attitudes and emotions in human-human interactions, which are characterised by complex and variable nature of interpersonal communications \cite{hess1998facial}. As a result, the development of automatic personalised facial reaction generation (FRG) solutions would enable interactive systems and humanoid virtual agents to express human-style facial reactions that better match the target individual's facial behaviour style, improving naturalness, engagement, and user trust compared to generic FRG systems.

Early FRG solutions \cite{huang2017dyadgan,yoon2022genea,song2022learning,ng2022learning,shao2021personality,zhou2022responsive} aim to reproduce a specific facial reaction that resembles the ground-truth (real) facial reaction for each input speaker behaviour. Such deterministic solutions fail to consider the non-deterministic nature of human facial reactions, where human could express different but appropriate facial reactions (AFRs) from an identical speaker behaviour in real-world human-human interactions under different contexts \cite{mehrabian1974approach}, i.e., FRG is a `one-to-many mapping' task \cite{song2023multiple}. To bridge this gap, our previous work have introduced a new theoretical framework on why and how multiple AFRs can be generated for responding to a given speaker behaviour \cite{song2023multiple}, and successfully organised three consecutive multiple appropriate facial reaction generation (MAFRG) challenges, i.e., REACT 2023 \cite{song2023react2023}), REACT 2024 \cite{song2024react}, and REACT 2025 \cite{song2025react}, leading to the development of many successful online and offline MAFRG methods \cite{xu2026reversible,luo2024reactface,hoque2023beamer,liang2023unifarn,yu2023leveraging,nguyen2024vector,liu2024one,nguyen2025latent,tran2024dim,hu2024robust,lv2025hierarchical,nguyen2024multiple,dam2024finite,xie2025smooth,luo2025reactdiff,huang2025online,huang2025multiple,mao2025scattering,wang2025explaining} in the past three years. A typical example is that Xu et al. \cite{xu2026reversible} address the ill-posed training problem caused by `one-to-many mapping' nature of FRG by reformulating FRG model training as \textbf{one input} speaker behaviour correspond to \textbf{one AFR distribution} summarising multiple AFRs.

Despite such progresses in MAFRG, most existing MAFRG solutions are developed to generate general AFRs without considering personalised aspect, i.e., human listeners of different internal disposition (e.g., personality) can express varied AFRs in response to the same speaker behaviour \cite{song2022learning,shao2021personality}. While human facial reactions are inherently shaped by internal traits such as personality and cognitive states, they focus on generic FRG modelling but overlook the role of individual differences. Besides, previous MAFRG challenge dataset rarely integrate neurophysiological signals, limiting the exploration of internal–external alignment between latent user states and observable facial reactions. 

Addressing both limitations motivates the design of the REACT 2026 challenge. Following previous REACT challenges, the REACT 2026  invites participants to attend offline and online generic MAFRG tasks, both requiring to generate multiple AFRs in response to each given speaker audio-visual behaviour, based on the MARS dataset introduced by REACT 2025 Challenge. In addition, this version not only innovatively introduces the \textbf{offline personalised MAFRG and online personalised MAFRG tasks} but also provides extra personality labels and EEG signals along with multi-modal human expressive behaviours. Similar to previous REACT challenge, each generated AFR sequence is represented by a face video and the corresponding multi-channel facial primitive time-series consisting of 15 action units (AUs), 8 facial expressions, as well as valence and arousal intensities, where the metrics defined in \cite{song2023multiple} are employed to evaluate four aspects of the submitted models in terms of generated AFRs: appropriateness, diversity, realism and synchrony. Participants are required to submit their developed model, checkpoints and well-explained source code, accompanied by a paper describing their proposed methodologies and the achieved results. Only contributions that meet the pre-determined requirements, terms and conditions \footnote{https://sites.google.com/view/react2026/home} are eligible for participation. The organisers do not engage in active participation themselves, but instead undertake a re-evaluation of the findings of the systems submitted to all sub-challenges. The ranking of the submitted models depend on two metrics: correlation (FRCorr) of the generated facial reaction primitives and facial reaction realism (FRRea) of the generated facial reaction video clips. The main contributions and novelties of this challenge are summarised as follows: 
\begin{itemize}

    \item Introducing the extended MARS Dataset with newly introduced Big-Five personality labels and EEG signals, together with the recorded multi-modal audio, visual, objectively-annotated ground-truth AFRs of each speaker behaviour.

    \item Providing state-of-the-art personalised MAFRG baselines for generating personalised multiple AFRs in response to each multi-modal input (i.e., a speaker audio-visual behaviour) in a dyadic interaction setting. This is different from the previous REACT 2023, 2024, and 2025 challenges that only require to develop generic MAFRG models.


    \item Providing generic and personalised MAFRG baselines that not only generate AFRs but also estimate neural responses by generating EEG responses from the input speaker audio-visual behaviours.
    
\end{itemize}


\begin{figure}[tb]
    \centering
    \includegraphics[width=1\columnwidth]{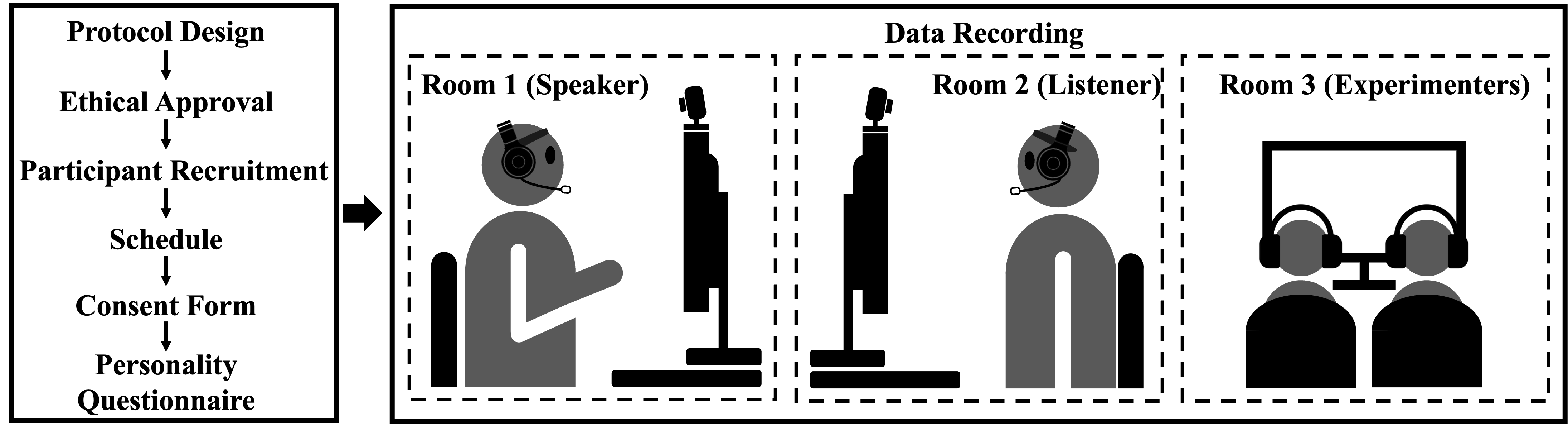}
    \caption{Illustration of the data collection of the MARS dataset. Left side outlines the preparatory steps, including protocol design, ethical approval, scheduling, obtaining participant consent, and completing a personality questionnaire. Right side illustrates the physical data collection setup, where a pair of human speaker and listener sit in front of PCs to conduct a video conference in the context of several pre-defined interaction tasks. Meanwhile, experimenters stay in the Room 3 to provide necessary helps and answers.}
    \label{fig:diagram}
\end{figure}



\begin{table}[h]
    \centering
    \caption{Statistics of different interaction topics of the MARS dataset. `s' denotes seconds and `m' denotes minutes.}
    \begin{adjustbox}{width=0.86\columnwidth,center}
        \begin{tabular}{lccc}
            \toprule
             \multirow{2}*{Topics} & \multicolumn{3}{c}{Time} \\
            \cline{2-4}
            & Min & Max & Average\\
            \midrule
            Cultural differences & 195s & 530s & 339s \\
            Movie scene sharing & 143s & 573s & 321s \\
            Policy changes & 153s & 726s & 336s \\
            Quizzes and games & 191s & 415s & 311s \\
            Scenario-based interviews & 161s & 518s & 305s \\
            \midrule
            Overall interactions & 19m'50s & 40m'49s & 27m'16s \\
            \bottomrule
        \end{tabular}
    \end{adjustbox}
    \label{tab:statistics}
\end{table}

\section{Challenge Tasks}

Given a short-term behaviour $b(S_n)^{t_1,t_2}$ expressed by a speaker $S_n$ at the time period $[t_1,t_2]$, the REACT 2026 challenge not only presents two generic Multiple Appropriate Facial Reaction Generation (MAFRG) tasks (i.e., offline and online generic MAFRG tasks), but also innovatively introduce two personalised MAFRG tasks (i.e., offline and online personalised MAFRG tasks).

\textbf{Task 1: Offline Generic MAFRG:} This task aims to develop a ML model $\mathcal{H}$ that takes the entire speaker behaviour sequence $b(S_n)^{t_1,t_2}$ as the input, and generates multiple ($M$) spatio-temporal AFRs 
$p_f(L \vert b(S_n)^{t_1,t_2})_1, \cdots, p_f(L \vert b(S_n)^{t_1,t_2})_M$, where $p_f(L \vert b(S_n)^{t_1,t_2})_m$ is a multi-channel time-series consisting of AUs, facial expressions, valence and arousal state representing the $m_\text{th}$ predicted facial reaction. As a result, $M$ AFRs are required to be generated for the task given each input speaker behaviour.

\textbf{Task 2: Online Generic MAFRG:} This task aims to develop a ML model $\mathcal{H}$ that estimates each frame (i.e.,  $\gamma_\text{th} \in [t_1,t_2]$ frame) of the target AFR by only considering the $\gamma_\text{th}$ frame and its previous frames expressed by the corresponding speaker (i.e., ${t_1}_\text{th}$ to $\gamma_\text{th}$ frames in $b(S_n)^{t}$), rather than taking all ${t_1}_\text{th}$ to ${t_2}_\text{th}$ frames into consideration. The model is expected to gradually generate all facial reaction frames to form multiple ($M$)  spatio-temporal AFRs $p_f(L \vert b(S_n)^{t_1,t_2})_1, \cdots, p_f(L \vert b(S_n)^{t_1,t_2})_M$, where $p_f(L \vert b(S_n)^{t_1,t_2})_m$ is a multi-channel facial attribute time-series (same as above) representing the $m_\text{th}$ predicted facial reaction. As a result, $M$ AFRs are required to be generated for the task given each input speaker behaviour.

\textbf{Task 3: Offline Personalised MAFRG:} This task aims to develop a ML model $\mathcal{H}$ that takes the entire speaker behaviour sequence $b(S_n)^{t_1,t_2}$, and generates multiple ($M$) personalised spatio-temporal AFRs $p_f(l \vert b(S_n)^{t_1,t_2})_1, \cdots, p_f(l \vert b(S_n)^{t_1,t_2})_M$, where \\
$p_f(l \vert b(S_n)^{t_1,t_2})_m$ is a multi-channel facial attribute time-series (same as above) representing the $m_\text{th}$ predicted personalised facial reaction, \textbf{which is expected to be similar to real AFRs expressed by the target listener $l$ in response to human behaviours that are similar to the given speaker behaviour $B^s$.} As a result, $M$ personalised AFRs are required to be generated for the task given each input speaker behaviour.

\textbf{Task 4: Online Personalised MAFRG:} This task aims to develop a ML model $\mathcal{H}$ that estimates each frame (i.e.,  $\gamma_\text{th} \in [t_1,t_2]$ frame) of the target AFR by only considering the $\gamma_\text{th}$ frame and its previous frames expressed by the corresponding speaker (i.e., ${t_1}_\text{th}$ to $\gamma_\text{th}$ frames in $b(S_n)^{t}$), rather than taking all ${t_1}_\text{th}$ to ${t_2}_\text{th}$ frames into consideration. The model is expected to gradually generate all facial reaction frames to form multiple ($M$)  spatio-temporal AFRs $p_f(L \vert b(S_n)^{t_1,t_2})_1, \cdots, p_f(L \vert b(S_n)^{t_1,t_2})_M$, where $p_f(L \vert b(S_n)^{t_1,t_2})_m$ is a multi-channel facial attribute time-series (same as above) representing the $m_\text{th}$ predicted personalised facial reaction, \textbf{which is expected to be similar to real AFRs expressed by the target listener $l$ in response to human behaviours that are similar to the given speaker behaviour $B^s$.} As a result, $M$ personalised AFRs are required to be generated for the task given each input speaker behaviour.

In addition, this challenge encourages participants to predict appropriate responsive brain activities (EEG signals) from given audio-visual speaker behaviours, with the ground-truth EEG signals labelled consistently with AFRs.

\section{Challenge Corpora}
\label{sec:dataset}

\textbf{Dataset.} The Multi-modal \textbf{M}ultiple \textbf{A}ppropriate \textbf{R}eaction in \textbf{S}ocial Dyads (MARS) Dataset \cite{song2025react} is the first audio, visual and EEG human behaviour MAFRG dataset (illustrated in Fig. \ref{fig:diagram}). It comprises 136 human-human dyadic interaction clips recorded from 23 human speakers and 136 human listeners, where each speaker is randomly and individually paired with 2 to 8 listeners. Specifically, each clip pair captures a human speaker and a human listener's behaviours, with each file containing 23 distinct sessions covering five main topics conducted in a fixed order, i.e., \textbf{cultural differences (four sessions per clip), movie scene sharing (four sessions per clip), policy changes (four sessions per clip), quizzes and games (eight sessions per clip), as well as scenario-based interviews (three sessions per clip)}. In total, 272 multi-modal recordings (136 clips) ranging from 19m50s to 40m49s are obtained. The statistics of the MARS dataset are provided in Table. \ref{tab:statistics}. Please refer to \cite{song2025react} for more details of the MARS dataset.

\textbf{Challenge data:} The REACT 2026 challenge dataset includes the following \textbf{raw and pre-processed data}: (1) original videos; (2) cropped face videos that still include expressions and head movements; (3) original audio signals; (5) textual transcripts; (6) EEG clips with 14 valid channels ("TP9", "AF7", "AF8", "TP10", "Delta\_TP9", "Theta\_TP9", "Alpha\_TP9", "Beta\_TP9", "Gamma\_TP9", "Delta\_TP10", "Theta\_TP10", "Alpha\_TP10", "Beta\_TP10", "Gamma\_TP10"); and (7) the age, gender, race, education level and the Big-Five personality traits of each participant. We also provide \textbf{frame-level facial and audio descriptors} including 15 AUs extracted using \cite{song2022gratis,luo2022learning}, eight facial expression probabilities and valence/arousal intensities extracted using EmoNet \cite{toisoul2021estimation}, 58-dimensional 3D Morphable Model (3DMM) coefficients (i.e., 52 facial expression coefficients, 3 pose coefficients and 3 translation coefficients) extracted using Faceverse \cite{wang2022faceverse} as well as 768-dimensional audio features extracted using wav2vec 2.0 \cite{baevski2020wav2vec}.

\textbf{Ground-truth labels.} During data recording, the semantic contexts are carefully controlled through 23 distinct sessions, where each is guided by a few pre-defined sentences posted by the speaker. This provides a consistent session-specific context across dyadic interactions between different speakers and listeners. In the context of \textbf{generic online and offline MAFRG tasks}, given each speaker behaviour expressed in a specific session, we define facial reactions expressed by all listeners under the same session to be AFRs (i.e., ground-truth) for responding to it. Alternatively, in the context of \textbf{personalised online and offline MAFRG tasks}, given each speaker behaviour expressed in a specific session, only the facial reaction expressed by the corresponding target listener in the same session is considered as the personalised AFR for responding to it. This way, the length of AFRs for responding to the same and different speaker behaviours can be varied.

\textbf{Ethical consideration.} Ethical approval for our MARS dataset was obtained at the University of Leicester. Prior to data acquisition, participants are required to read and sign the consent form. Subsequently, they complete the personality questionnaire and are informed about the recording equipments and their use. Finally, all participants are compensated with Amazon vouchers for their participation in the study. During the challenge, all participating teams are required to sign an EULA to access the dataset.

\section{Evaluation Protocol}

\noindent In this challenge, the submitted models are expected to generate two types of outputs for representing each AFR: a 25-channel facial attribute time-series and an AFR video. Since the lengths of real AFRs expressed by human listeners for responding to different or even same speaker behaviours can vary a lot, we train a Transformer-based variational encoder-decoder framework to summarise every arbitrary-length real AFRs and generated AFRs into a fixed-length sequence of tokens, representing them by token sequences of the same length in a latent space. This not only facilitates developing MAFRG models that can process variable-length input speaker behaviours (represented by fixed-length token sequences) and accordingly generate fixed-length AFR representations, but also allow the generated AFRs to be directly compared with all corresponding real AFRs of varied lengths in this latent space. Then, a transformer decoder is also developed to decode the variable-length token sequences as AFR videos of the required length.

For \textbf{generic MAFRG tasks}, we follow \cite{song2023multiple,song2023react2023} to comprehensively evaluate three aspects of the generated facial reaction attributes: (1) \textbf{Appropriateness} based on two metrics, \textbf{FRCorr}: Concordance Correlation Coefficient (CCC) and \textbf{FRDist}: Dynamic Time Warping (DTW); (2) \textbf{Diversity}: \textbf{FRVar}, and \textbf{FRDiv}; and (3) \textbf{Synchrony}: the Time Lagged Cross Correlation (TLCC), called \textbf{FRSyn} in this challenge. Also, the \textbf{Realism} of the generated facial reaction video clips is assessed using the Fréchet Inception Distance (FID), denoted as \textbf{FRRea}.

For \textbf{personalised MAFRG tasks}, we also compute the above metrics based on the same types of outputs. The only difference is that the personalised MAFRG evaluation only assigns the GT real AFR expressed by the corresponding conversational partner, i.e., i.e., the listener of the target speaker, rather than defining multiple real AFRs expressed by different listeners as the ground-truths (GT) for each speaker behaviour.



\begin{figure*}[tb!]
    \centering
    \includegraphics[width=1.0\linewidth]{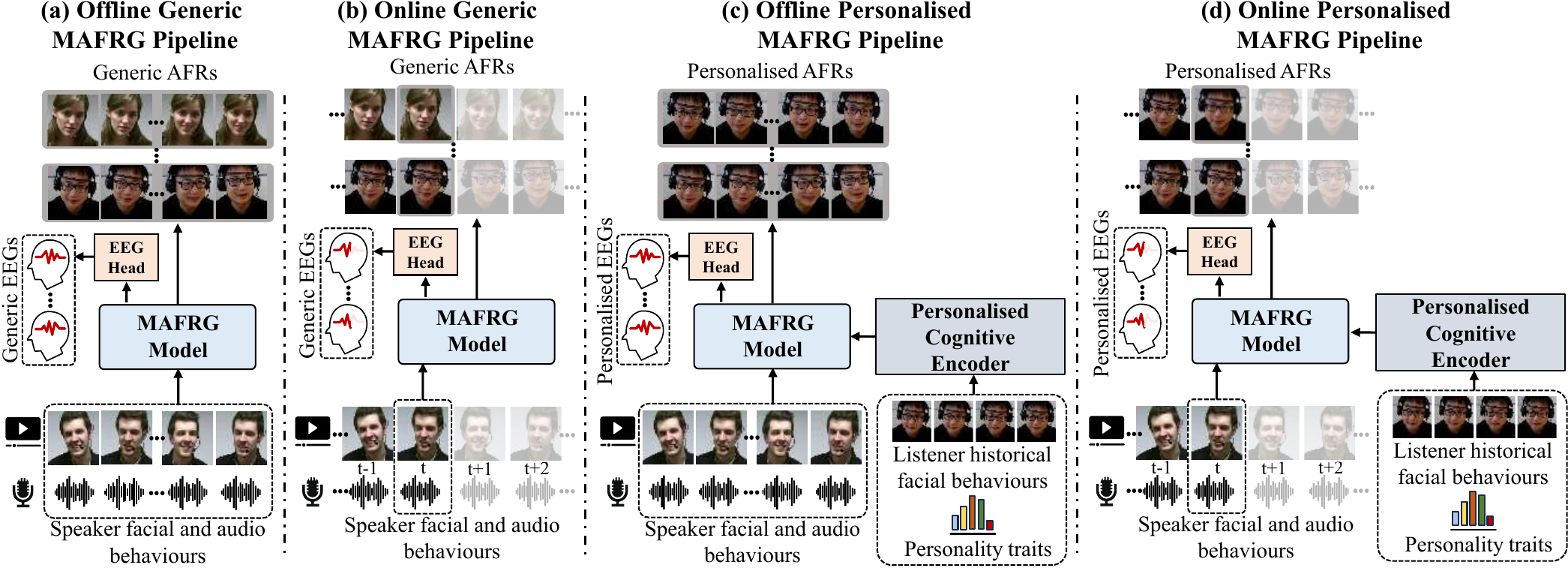}
    \caption{Pipelines for four target MAFRG tasks. \textbf{Offline generic MAFRG models} directly takes each entire speaker audio-visual behaviour sequence as the input to generate the corresponding AFRs sequences/videos. \textbf{Online generic MAFRG models} consider the short speaker behaviour segment at the current time step, which outputs AFR segments conditioned on it and previous speaker behaviours in an ongoing manner. \textbf{Personalised MAFRG models} additionally encode personalised factors of the target listener to personalise the output of the generic MAFRG models.}
    \label{fig:main}
\end{figure*}


\section{Baseline Models}

\noindent We follow REACT 2025 \cite{song2025react} to employ the open-source Trans-VAE \cite{song2023react2023,song2024react,song2025react,luo2024reactface} and PerFRDiff \cite{zhu2024perfrdiff} as the online and offline generic MAFRG baselines, while using the Reversible Graph Neural Network (REGNN)  \cite{xu2026reversible} as the offline generic MAFRG baseline. Different from REACT 2025, we have additionally added a head consisting of a 2-layer MLP for predicting multiple appropriate EEG responses in addition to AFRs. Meanwhile, we employ PerFRDiff \cite{zhu2024perfrdiff} for the online personalised MAFRG baseline and extend a simple version of PerReactor \cite{zhu2025perreactor} (S-PerReactor) as the offline personalised MAFRG baseline, where we also use a 2-layer MLP head for predicting appropriate EEG responses. Fig. \ref{fig:main} illustrates how the developed baselines achieved the four sub-tasks introduced by this challenge.

\textbf{Trans-VAE} 
is made up of: (i) a \textbf{CNN encoder} extracting facial reaction-related features from the input speaker facial image sequence; (ii) a \textbf{Transformer encoder} combining the learned facial and audio features (extracted by wav2vec 2.0 \cite{baevski2020wav2vec}), based on which a Gaussian distribution is learned to describe multiple predicted AFRs; and (iii) a \textbf{Transformer decoder} generating two facial reaction descriptors from the distribution: a set of 3DMM coefficients \cite{wang2022faceverse}
and a 25-channel facial attribute time-series (Refer to \cite{song2023react2023} for details), as well as EEG responses. 




\textbf{REGNN} consists of three main modules: (i) a \textbf{Perceptual Processor} that encodes the input speaker audio-facial behaviour as a pair of audio and facial representations; (ii) a \textbf{Cognitive Processor} that predicts a Gaussian Mixture Graph Distribution describing all AFRs of the input speaker behaviour; and (iii) a reversible GNN-based \textbf{Motor Processor} that samples multiple AFRs from the predicted distribution.

\textbf{PerFRDiff} employs three modality-specific linear projectors to individually map the input speaker audio features extracted from Wav2vec 2.0 \cite{baevski2020wav2vec}, multi-channel facial attribute time-series, and 3DMM coefficients sequence, to a shared latent space. These are input as conditions to a diffusion Transformer for jointly denoising AFRs and EEG responses. We follow REACT 2025 to re-train its generic MAFRG part to separately conduct online and offline generic MAFRG tasks. The online personalised MARFG is achieved by applying its original version, where a personalised behaviour encoder learns a set of weight shifts from the target listener's historical facial behaviours, editing the generic diffusion-based MAFRG model as a personalised MAFRG model.

\textbf{S-PerReactor} is a Generative Adversarial Network-style model. In its generator, a Transformer first learns a generic AFR distribution from the given speaker audio-visual behaviour, while a CNN-attention network extracts personalised behaviour patterns from the target listener’s historical facial video to generate a personalised facial behaviour condition. This condition is also fed to the main branch of the generator to enforce it outputting personalised AFRs.

\section{Baseline Results}

\begin{table}[t]
  \Huge
  \centering
  \caption{MAFRG baseline results achieved on the MARS test set, where B\_Random randomly samples $\alpha = 10$ AFR sequences from a Gaussian distribution; B\_Mime generates AFRs by mimicking the corresponding speaker facial behaviours; and B\_MeanFr generates AFRs by averaging the frame-wise AFRs in the training set, respectively. LHFB denotes the 3DMM time-series of the target listener's historical facial behaviour.}
  \label{tab:baseline-results-face}
  \begin{adjustbox}{width=1\columnwidth}
\begin{tabular}{lcccccccc}
\toprule
  &
  \multicolumn{2}{c}{Personalised condition} &
  \multicolumn{2}{c}{Appropriateness} &
  \multicolumn{2}{c}{Diversity} &
  Realism &
  Synchrony \\ 
  \cmidrule(lr){2-3} \cmidrule(lr){4-5} \cmidrule(lr){6-7} \cmidrule(lr){8-8} \cmidrule(lr){9-9}
  \multirow{-2}{*}{Method} &
  LHFB &
  Personality &
  FRCorr (↑) &
  FRDist (↓) &
  FRDiv (↑) &
  FRVar (↑) &
  FRRea (↓) &
  FRSyn (↓) \\ \midrule
GT & & & 10 & 0.00 & 0.1926 & 0.0623 & -- & 48.80 \\
B\_Random & & & 0.0360 & 432.14 & 0.3343 & 0.1671 & --  & 47.62 \\
B\_Mime & & & 0.5454 & 180.68 & 0.0000 & 0.0756 & --  & 45.79 \\
B\_MeanFr & & & 0.0000 & 163.61 & 0.0000 & 0.0000 & -- & 49.00 \\
\midrule
\multicolumn{7}{l}{\bfseries Offline generic MAFRG results} \\
\midrule
Trans-VAE & & & 0.2582 & 157.57 & 0.0078 & 0.0069 & - & 49.00 \\
REGNN & & & 0.4770 & 130.30 & 0.0014 & 0.0040 & - & 46.22 \\
\midrule
\multicolumn{7}{l}{\bfseries Online generic MAFRG results} \\
\midrule
Trans-VAE & & & 0.3483 & 182.46 & 0.0893 & 0.0500 & - & 49.00 \\
PerFRDiff & & & 0.5680 & 199.56 & 0.1503 & 0.0867 & - & 47.65 \\
\midrule
\multicolumn{7}{l}{\bfseries Offline personalised MAFRG results} \\
\midrule
 & $\checkmark$ &           & - & - & - & - & -  & - \\
 &           & $\checkmark$ & - & - & - & - & -  & - \\
\multirow{-3}{*}{S-PerReactor} 
& $\checkmark$ & $\checkmark$ & - & - & - & - & -  & - \\
\midrule
\multicolumn{7}{l}{\bfseries Online personalised MAFRG results} \\
\midrule
 & $\checkmark$ &           & 0.5497 & 204.36 & 0.1515 & 0.0872 & - & 47.88 \\
 &           & $\checkmark$ & 0.5475 & 196.37 & 0.1452 & 0.0832 & - & 47.40 \\
\multirow{-3}{*}{PerFRDiff} 
& $\checkmark$ & $\checkmark$ & 0.5453 & 192.54 & 0.1445 & 0.0828 & - & 47.69 \\ \bottomrule
\end{tabular}
\end{adjustbox}
\label{tab:MAFRG-results}
\end{table}

\begin{table}[t]
  \Huge
  \centering
  \caption{Appropriate EEG reaction generation baseline results achieved on the MARS test set.}
  \label{tab:baseline-results-eeg}
  \begin{adjustbox}{width=1\columnwidth}
\begin{tabular}{lcccccccc}
\toprule
  &
  \multicolumn{2}{c}{Input(s)} &
  \multicolumn{2}{c}{Appropriateness} &
  \multicolumn{2}{c}{Diversity} \\ 
  \cmidrule(lr){2-3} \cmidrule(lr){4-5} \cmidrule(lr){6-7} 
  \multirow{-2}{*}{Method} &
  LHFB &
  Personality &
  EEGCorr (↑) &
  EEGDist (↓) &
  EEGDiv (↑) &
  EEGVar (↑) & 
  \\ \midrule
\multicolumn{7}{l}{\bfseries Offline Generic Results} \\
\midrule
Trans-VAE & & & 0.1719 & 5.73 & 0.0000 & 0.0012   \\
REGNN & & & 0.0671 & 19.42 & 0.0000 & 0.0025   \\
\midrule
\multicolumn{7}{l}{\bfseries Online Generic Results} \\
\midrule
Trans-VAE & & & 0.1362 & 9.28 & 0.0001 & 0.0020   \\
PerFRDiff & & & 0.0731 & 21.15 & 0.0047 & 0.0079   \\
\midrule
\multicolumn{7}{l}{\bfseries Offline personalised MAFRG results} \\
\midrule
 & $\checkmark$ &           & - & - & - & - \\
 &           & $\checkmark$ & - & - & - & -  \\
\multirow{-3}{*}{S-PerReactor} 
& $\checkmark$ & $\checkmark$ & - & - & - & -  \\
\multicolumn{7}{l}{\bfseries Online Personalised Results} \\
\midrule
 & $\checkmark$ &           & 0.0722 & 22.75 & 0.0052 & 0.0072   \\
 &           & $\checkmark$ & 0.0893 & 22.67 & 0.0043 & 0.0067   \\
\multirow{-3}{*}{PerFRDiff} 
& $\checkmark$ & $\checkmark$ & 0.0747 & 22.69 & 0.0045 & 0.0068  \\ \bottomrule
\end{tabular}
\end{adjustbox}
\label{tab:EEG-results}
\end{table}

\textbf{Generic MAFRG task:} Table \ref{tab:baseline-results-face} show that all generic MAFRG baselines clearly outperformed the B\_Random and B\_MeanFr, which validated that they can to some extent generate meaningful AFRs from different speaker behaviours. Here, the strong results of $B\_Mime$ suggest that human listeners may frequently exhibit facial behaviours that mirror those of the corresponding speakers, i.e., facial mimicry \cite{dimberg1982facial}. From another perspective, the diffusion-based PerFRDiff baseline outperformed the Trans-VAE for online MAFRG tasks by generating more correlated, diverse and synchronised AFRs, while REGNN generated more appropriate and synchronised AFRs over the Trans-VAE under the offline setting. Both results suggest that the distribution learning strategy proposed by REGNN and the diffusion algorithm of PerFRDiff are more powerful than Trans-VAE in modelling the one-to-many mapping nature between speaker behaviours and human facial reactions.

\textbf{Personalised MAFRG task:} In terms of the personalised MAFRG tasks, Table \ref{tab:baseline-results-face} indicates that the AFRs generated by the PerFRDiff baseline show clear positive association with the GT real personalised AFRs expressed by the corresponding target individuals. It is reasonable that personalised models do not necessarily achieve better appropriateness performance, measured by FRCorr and FRDist, than their generic variants. For each generated AFR, the evaluation compares it with the corresponding set of real AFRs and selects the most similar one for metric computation. In the case of personalised models, this comparison is restricted to the real AFRs expressed by the target listener only. By contrast, AFRs generated by generic models are compared against real AFRs from all listeners, which naturally includes those expressed by the target listener.

\textbf{EEG reaction prediction task:} Table \ref{tab:baseline-results-eeg}  reports appropriate EEG reaction prediction results, which provided the first evidence that human listeners' brain activities can be partially predicted from only speaker audio-visual expressive behaviours. This is because baseline under all conditions output predictions that are positively related with the GT appropriate EEGs despite they are weakly associated.

\section{Participation and Conclusion}

This paper introduces REACT 2026 Challenge in conjunction with the ACM Multimedia 26 Conference, which focuses on generic and personalised MAFRG tasks under various video conference-based dyadic interactions scenarios. 
Challenge participants were given access to the training and validation sets to develop their ML models, with a challenge guideline and a baseline paper released to provide more details. Each participating group was allowed to submit their models and results for the test set, which were objectively evaluated. Our evaluation protocol strictly ranked all participant models under the same settings by evaluating two aspects of their generated facial reactions: FRCorr and FRRea. We hope that the new personalised baseline code, methodologies and results of the participating teams, will serve as a valuable stepping stone for researchers and practitioners interested in automatic facial reaction generation.

\bibliographystyle{ACM-Reference-Format}
\bibliography{bibliography}

@article{song2024react,
  title={React 2024: the second multiple appropriate facial reaction generation challenge},
  author={Song, Siyang and Spitale, Micol and Luo, Cheng and Palmero, Cristina and Barquero, German and Zhu, Hengde and Escalera, Sergio and Valstar, Michel and Baur, Tobias and Ringeval, Fabien and others},
  journal={arXiv preprint arXiv:2401.05166},
  year={2024}
}

@inproceedings{huang2017dyadgan,
  title={Dyadgan: Generating facial expressions in dyadic interactions},
  author={Yuchi et al},
  booktitle={IEEE CVPR Workshops 2017},
  pages={11--18},
  year={2017}
}

@inproceedings{shao2021personality,
  title={Personality recognition by modelling person-specific cognitive processes using graph representation},
  author={Shao, Zilong and Song, Siyang and Jaiswal, Shashank and Shen, Linlin and Valstar, Michel and Gunes, Hatice},
  booktitle={proceedings of the 29th ACM international conference on multimedia},
  pages={357--366},
  year={2021}
}

@inproceedings{nguyen2025latent,
  title={Latent behaviour Diffusion for Sequential Reaction Generation in Dyadic Setting},
  author={Nguyen, Minh-Duc and Yang, Hyung-Jeong and Kim, Soo-Hyung and Shin, Ji-Eun and Kim, Seung-Won},
  booktitle={International Conference on Pattern Recognition},
  pages={233--248},
  year={2025},
  organization={Springer}
}

@article{xu2026reversible,
  title={Reversible graph neural network-based reaction distribution learning for multiple appropriate facial reactions generation},
  author={Xu, Tong and Spitale, Micol and Tang, Hao and Liu, Lu and Gunes, Hatice and Song, Siyang},
  journal={IEEE Transactions on Affective Computing},
  year={2026},
  publisher={IEEE}
}

@inproceedings{mao2025scattering,
  title={Scattering-Conditioned Diffusion Models for Multiple Appropriate Facial Reaction Generation},
  author={Mao, Qirong and Wu, Qiwei and Liu, Na and Ding, Yakui and Gao, Lijian},
  booktitle={Proceedings of the 33rd ACM International Conference on Multimedia},
  pages={13985--13991},
  year={2025}
}

@inproceedings{huang2025multiple,
  title={Multiple Appropriate Facial Reaction Generation Based on Multi-View Transformation of Speaker Video},
  author={Huang, Jiajian and Yu, Zitong},
  booktitle={Proceedings of the 33rd ACM International Conference on Multimedia},
  pages={13992--13996},
  year={2025}
}

@inproceedings{huang2025online,
  title={Online Emotion-Driven Generation of Multiple Appropriate Facial Reactions},
  author={Huang, Jiajian and Song, Siyang and Kong, Xiangyu and Xie, Weicheng and Shen, Linlin and Yu, Zitong},
  booktitle={Chinese Conference on Biometric Recognition},
  pages={183--194},
  year={2025},
  organization={Springer}
}

@inproceedings{luo2025reactdiff,
  title={ReactDiff: Fundamental Multiple Appropriate Facial Reaction Diffusion Model},
  author={Luo, Cheng and Song, Siyang and Yan, Siyuan and Yu, Zhen and Ge, Zongyuan},
  booktitle={Proceedings of the 33rd ACM International Conference on Multimedia},
  pages={5607--5616},
  year={2025}
}

@inproceedings{song2025react,
  title={React 2025: the third multiple appropriate facial reaction generation challenge},
  author={Song, Siyang and Spitale, Micol and Kong, Xiangyu and Zhu, Hengde and Luo, Cheng and Palmero, Cristina and Barquero, German and Escalera, Sergio and Valstar, Michel and Daoudi, Mohamed and others},
  booktitle={Proceedings of the 33rd ACM International Conference on Multimedia},
  pages={13979--13984},
  year={2025}
}

@inproceedings{xie2025smooth,
  title={Smooth Online Multiple Appropriate Facial Reaction Generation},
  author={Xie, Weicheng and Yan, Chunlin and Song, Siyang and Yu, Zitong and Shen, Linlin and Cui, Laizhong},
  booktitle={Proceedings of the 33rd ACM International Conference on Multimedia},
  pages={5804--5813},
  year={2025}
}

@inproceedings{wang2025explaining,
  title={Explaining Listener Reactions: Personality-Guided Facial Response Generation with Cross-Modal Attention},
  author={Wang, Peng and Xue, Pujun and Liu, Xiaofeng and Ji, Tongjuan},
  booktitle={Proceedings of the 33rd ACM International Conference on Multimedia},
  pages={13997--14003},
  year={2025}
}

@inproceedings{zhu2024perfrdiff,
  title={Perfrdiff: Personalised weight editing for multiple appropriate facial reaction generation},
  author={Zhu, Hengde and Kong, Xiangyu and Xie, Weicheng and Huang, Xin and Shen, Linlin and Liu, Lu and Gunes, Hatice and Song, Siyang},
  booktitle={Proceedings of the 32nd ACM International Conference on Multimedia},
  pages={9495--9504},
  year={2024}
}

@inproceedings{zhu2025perreactor,
  title={PerReactor: Offline Personalised Multiple Appropriate Facial Reaction Generation},
  author={Zhu, Hengde and Kong, Xiangyu and Xie, Weicheng and Huang, Xin and He, Xilin and Liu, Lu and Shen, Linlin and Zhang, Wei and Gunes, Hatice and Song, Siyang},
  booktitle={Proceedings of the AAAI Conference on Artificial Intelligence},
  volume={39},
  number={2},
  pages={1665--1673},
  year={2025}
}

@inproceedings{lv2025hierarchical,
  title={Hierarchical Multimodal Decoupling-Fusion Framework for offline Multiple Appropriate Facial Reaction Generation},
  author={Lv, Qincheng and Liu, Xiaofeng and Li, Jie and Ni, Rongrong and Xue, Pujun and Song, Siyang},
  booktitle={ICASSP 2025-2025 IEEE International Conference on Acoustics, Speech and Signal Processing (ICASSP)},
  pages={1--5},
  year={2025},
  organization={IEEE}
}

@inproceedings{hu2024robust,
  title={Robust Facial Reactions Generation: An Emotion-Aware Framework with Modality Compensation},
  author={Hu, Guanyu and Wei, Jie and Song, Siyang and Kollias, Dimitrios and Yang, Xinyu and Sun, Zhonglin and Kaloidas, Odysseus},
  booktitle={2024 IEEE International Joint Conference on Biometrics (IJCB)},
  pages={1--10},
  year={2024},
  organization={IEEE}
}

@inproceedings{tran2024dim,
  title={DIM: Dyadic Interaction Modeling for Social Behavior Generation},
  author={Tran, Minh and Chang, Di and Siniukov, Maksim and Soleymani, Mohammad},
  booktitle={European Conference on Computer Vision},
  pages={484--503},
  year={2024},
  organization={Springer}
}

@article{song2022learning,
  title={Learning Person-specific Cognition from Facial Reactions for Automatic Personality Recognition},
  author={Song et al},
  journal={IEEE Transactions on Affective Computing},
  year={2022},
  publisher={IEEE}
}

@book{mehrabian1974approach,
  title={An approach to environmental psychology.},
  author={Mehrabian, Albert and Russell, James A},
  year={1974},
  publisher={the MIT Press}
}

@article{yoon2022genea,
  title={The GENEA Challenge 2022: A large evaluation of data-driven co-speech gesture generation},
  author={Yoon et al},
  journal={arXiv preprint arXiv:2208.10441},
  year={2022}
}

@article{toisoul2021estimation,
  title={Estimation of continuous valence and arousal levels from faces in naturalistic conditions},
  author={Toisoul, Antoine and et al},
  journal={Nature Machine Intelligence},
  volume={3},
  number={1},
  pages={42--50},
  year={2021},
  publisher={Nature Publishing Group}
}

@article{luo2022learning,
  title={Learning Multi-dimensional Edge Feature-based AU Relation Graph for Facial Action Unit Recognition},
  author={Luo, Cheng and et al},
  journal={arXiv preprint arXiv:2205.01782},
  year={2022}
}

@article{song2023multiple,
  title={Multiple Appropriate Facial Reaction Generation in Dyadic Interaction Settings: What, Why and How?},
  author={Song et al},
  journal={https://arxiv.org/abs/2302.06514},
  year={2023}
}

@inproceedings{yu2023leveraging,
  title={Leveraging the Latent Diffusion Models for Offline Facial Multiple Appropriate Reactions Generation},
  author={Yu, Jun and Zhao, Ji and Xie, Guochen and Chen, Fengxin and Yu, Ye and Peng, Liang and Li, Minglei and Dai, Zonghong},
  booktitle={Proceedings of the ACM International Conference on Multimedia},
  pages={9561--9565},
  year={2023}
}

@article{luo2024reactface,
  title={ReactFace: Online Multiple Appropriate Facial Reaction Generation in Dyadic Interactions},
  author={Luo, Cheng and Song, Siyang and Xie, Weicheng and Spitale, Micol and Ge, Zongyuan and Shen, Linlin and Gunes, Hatice},
  journal={IEEE Transactions on Visualization and Computer Graphics},
  year={2024},
  publisher={IEEE}
}

@inproceedings{hoque2023beamer,
  title={BEAMER: Behavioral Encoder to Generate Multiple Appropriate Facial Reactions},
  author={Hoque, Ximi and Mann, Adamay and Sharma, Gulshan and Dhall, Abhinav},
  booktitle={Proceedings of the ACM International Conference on Multimedia},
  pages={9536--9540},
  year={2023}
}

@article{hess1998facial,
  title={Facial reactions to emotional facial expressions: Affect or cognition?},
  author={Hess, Ursula and Phillippot, Pierre and Blairy, Sylvie},
  journal={Cognition \& Emotion},
  volume={12},
  number={4},
  pages={509--531},
  year={1998}
}

@article{dimberg1982facial,
  title={Facial reactions to facial expressions},
  author={Dimberg, Ulf},
  journal={Psychophysiology},
  volume={19},
  number={6},
  pages={643--647},
  year={1982},
  publisher={Wiley Online Library}
}

@inproceedings{zhou2022responsive,
  title={Responsive listening head generation: a benchmark dataset and baseline},
  author={Zhou, Mohan and Bai, Yalong and Zhang, Wei and Yao, Ting and Zhao, Tiejun and Mei, Tao},
  booktitle={European conference on computer vision},
  pages={124--142},
  year={2022},
  organization={Springer}
}

@inproceedings{liang2023unifarn,
  title={UniFaRN: Unified Transformer for Facial Reaction Generation},
  author={Liang, Cong and Wang, Jiahe and Zhang, Haofan and Tang, Bing and Huang, Junshan and Wang, Shangfei and Chen, Xiaoping},
  booktitle={Proceedings of the ACM International Conference on Multimedia},
  pages={9506--9510},
  year={2023}
}

@inproceedings{dam2024finite,
  title={Finite Scalar Quantization as Facial Tokenizer for Dyadic Reaction Generation},
  author={Dam, Quang Tien and Nguyen, Tri Tung Nguyen and Tran, Dinh Tuan and Lee, Joo-Ho},
  booktitle={2024 IEEE 18th International Conference on Automatic Face and Gesture Recognition (FG)},
  pages={1--5},
  year={2024},
  organization={IEEE}
}

@inproceedings{liu2024one,
  title={One-to-Many Appropriate Reaction Mapping Modeling with Discrete Latent Variable},
  author={Liu, Zhenjie and Liang, Cong and Wang, Jiahe and Zhang, Haofan and Liu, Yadong and Zhang, Caichao and Gui, Jialin and Wang, Shangfei},
  booktitle={2024 IEEE 18th International Conference on Automatic Face and Gesture Recognition (FG)},
  pages={1--5},
  year={2024},
  organization={IEEE}
}

@inproceedings{nguyen2024vector,
  title={Vector Quantized Diffusion Models for Multiple Appropriate Reactions Generation},
  author={Nguyen, Minh-Duc and Yang, Hyung-Jeong and Ho, Ngoc-Huynh and Kim, Soo-Hyung and Kim, Seungwon and Shin, Ji-Eun},
  booktitle={2024 IEEE 18th International Conference on Automatic Face and Gesture Recognition (FG)},
  pages={1--5},
  year={2024},
  organization={IEEE}
}

@inproceedings{nguyen2024multiple,
  title={Multiple Facial Reaction Generation Using Gaussian Mixture of Models and Multimodal Bottleneck Transformer},
  author={Nguyen, Dang-Khanh and Paudel, Prabesh and Kim, Seung-Won and Shin, Ji-Eun and Kim, Soo-Hyung and Yang, Hyung-Jeong},
  booktitle={2024 IEEE 18th International Conference on Automatic Face and Gesture Recognition (FG)},
  pages={1--5},
  year={2024},
  organization={IEEE}
}

@inproceedings{song2023react2023,
  title={React2023: The first multiple appropriate facial reaction generation challenge},
  author={Song, Siyang and Spitale, Micol and Luo, Cheng and Barquero, Germ{\'a}n and Palmero, Cristina and Escalera, Sergio and Valstar, Michel and Baur, Tobias and Ringeval, Fabien and Andr{\'e}, Elisabeth and others},
  booktitle={Proceedings of the 31st ACM International Conference on Multimedia},
  pages={9620--9624},
  year={2023}
}

@inproceedings{ng2022learning,
  title={Learning to listen: Modeling non-deterministic dyadic facial motion},
  author={Evonne Ng et al},
  booktitle={IEEE/CVF CVPR 2022},
  pages={20395--20405},
  year={2022}
}

@article{baevski2020wav2vec,
  title={wav2vec 2.0: A framework for self-supervised learning of speech representations},
  author={Baevski, Alexei and Zhou, Yuhao and Mohamed, Abdelrahman and Auli, Michael},
  journal={Advances in neural information processing systems},
  volume={33},
  pages={12449--12460},
  year={2020}
}

@article{song2022gratis,
  title={Gratis: Deep learning graph representation with task-specific topology and multi-dimensional edge features},
  author={Song, Siyang and Song, Yuxin and Luo, Cheng and Song, Zhiyuan and Kuzucu, Selim and Jia, Xi and Guo, Zhijiang and Xie, Weicheng and Shen, Linlin and Gunes, Hatice},
  journal={arXiv preprint arXiv:2211.12482},
  year={2022}
}

@inproceedings{wang2022faceverse,
  title={Faceverse: a fine-grained and detail-controllable 3d face morphable model from a hybrid dataset},
  author={Wang, Lizhen and Chen, Zhiyuan and Yu, Tao and Ma, Chenguang and Li, Liang and Liu, Yebin},
  booktitle={Proceedings of the IEEE/CVF conference on computer vision and pattern recognition},
  pages={20333--20342},
  year={2022}
}

\end{document}